\documentclass[letterpaper, 10 pt, conference]{ieeeconf}  

\IEEEoverridecommandlockouts                              

\usepackage{graphicx}
\graphicspath{{images/}}
\usepackage{adjustbox}
\usepackage{multirow}
\usepackage{amssymb}
\usepackage{color}
\usepackage{tikz}
\usepackage{pgfplotstable}
\usepackage{pgfplots}
\usepackage[normalem]{ulem}

\usepackage{cite}

\usepackage{lineno}
\usepackage{subfig}
\usepackage{tabularx, rotating, caption}
\usepackage{array}
\usepackage{siunitx}
\usepackage{url}
\usepackage{booktabs}
\usepackage{comment}
\usepackage{cleveref}
\usepackage{pifont}
\usepackage{lipsum}

\newcommand \red[1]{#1}

\title{\Large \bf 
Estimating Map Completeness in Robot Exploration }

\author{Matteo Luperto$^{1}$, Marco Maria Ferrara$^{2}$, Giacomo Boracchi$^{2}$,  and Francesco Amigoni$^{2}$
\thanks{$^{1}$M. Luperto is with the Dipartimento di Informatica of the Università degli Studi di Milano, Italy
        {\tt\small matteo.luperto@unimi.it}}
\thanks{$^{2}$M. Ferrara, G. Boracchi, and F. Amigoni are with the Dipartimento di Elettronica, Informazione e Bioingegneria of the Politecnico di Milano, Italy
        {\tt\small marcomaria.ferrara@mail.polimi.it, \{giacomo.boracchi,francesco.amigoni\}@polimi.it}}
}

\begin{document}

\maketitle
\thispagestyle{empty}
\pagestyle{empty}

\begin{abstract}
In this paper, we propose a method that, given a partial grid map of an indoor environment built by an autonomous mobile robot, estimates the amount of the explored area represented in the map, as well as whether the uncovered part is still worth being explored or not.
Our method is based on a deep convolutional neural network trained on data from partially explored environments with annotations derived from the knowledge of the entire map (which is not available when the network is used for inference). We show how such a network can be used to define a stopping criterion to terminate the exploration process when it is no longer adding relevant details about the environment to the map, saving, on average, $40\%$ of the total exploration time with respect to covering all the area of the environment.
\end{abstract}

\section{Introduction}\label{sec:intro}

In exploration for map building, an autonomous mobile robot builds a representation, or \textit{map}, of an initially unknown indoor environment by iteratively performing a sequence of steps~\cite{Thrun02a}. 
First, the robot identifies a set of reachable candidate locations within the known portion of the environment represented by the current map. Usually, these candidate locations are at the boundaries, called \textit{frontiers}, between known and unknown parts of the environment. The robot then selects the most promising location to reach to progress exploration, according to an exploration strategy. Finally, while reaching the selected location, the robot updates its map using newly acquired sensor data. Then, the cycle repeats.


\begin{figure}[t]
\centering
\subfloat[\texttt{$t=\SI{90}{\minute}$}.\label{fig:map30percent}]{\includegraphics[trim={0 4cm 0 4cm},clip,width=0.85\linewidth]{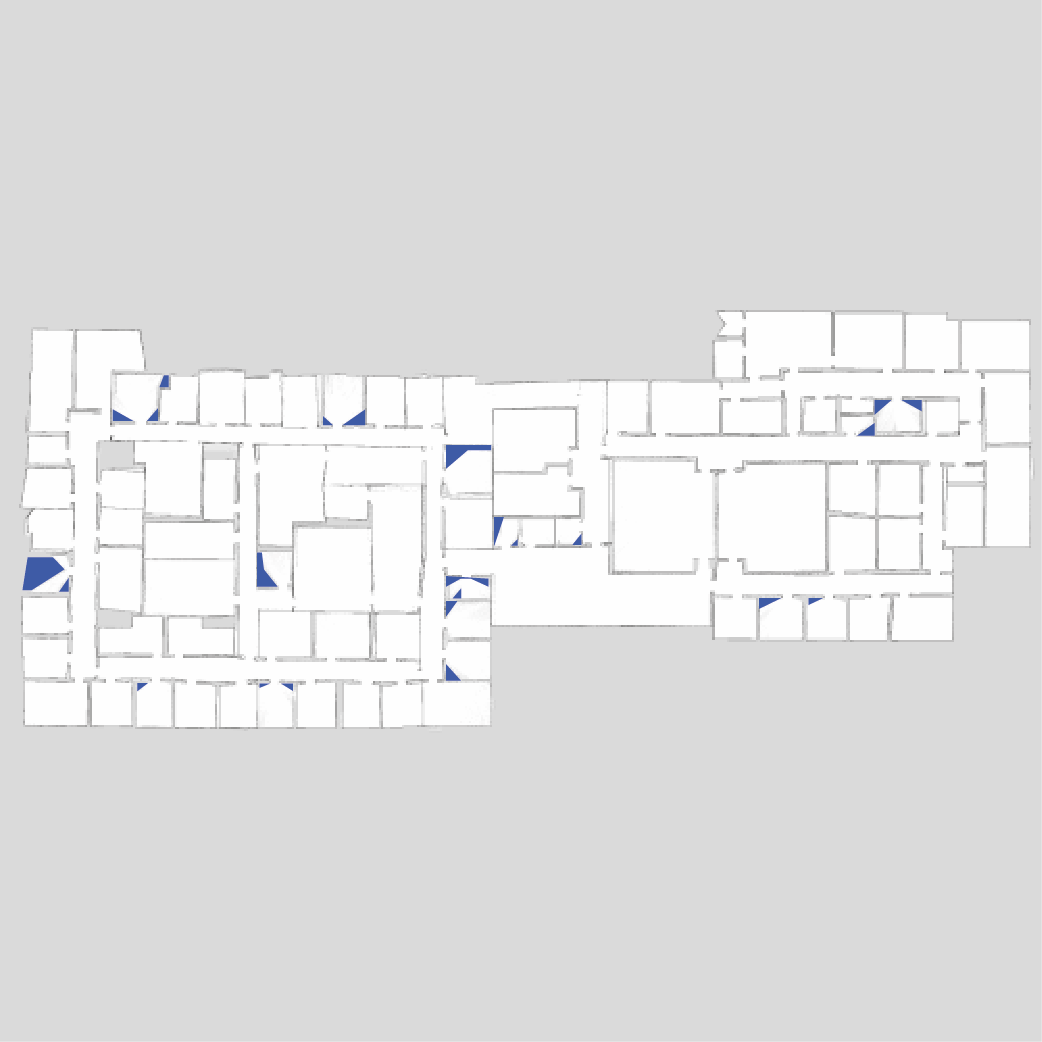}}\\
\subfloat[\texttt{$T=\SI{150}{\minute}$}.\label{fig:map100percent}]{\includegraphics[trim={0 4cm 0 4cm},clip,width=0.85\linewidth]{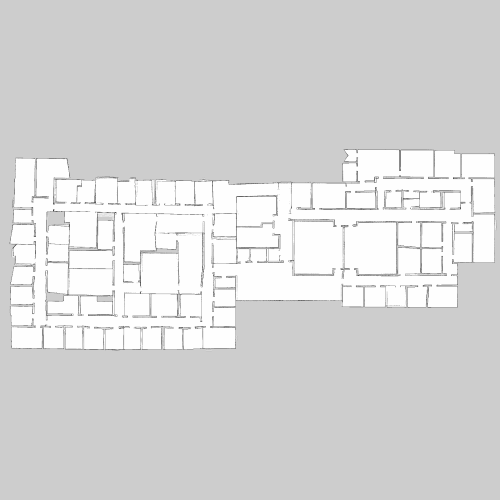}}
\caption{
The map after \texttt{$t=\SI{90}{\minute}$} of exploration (a) already represents almost all of the environment (the portion of the mapped area is $A_{t}=0.98$), except a few uninteresting corners and portions of rooms scattered across the whole environment, highlighted in blue in (a). This exploration run ends at $T=\SI{150}{\minute}$ (b) when all frontiers have been visited. During the time interval $T-t$, the robot moves back and forth to explore the remaining frontiers, possibly jeopardizing the whole mapping process.
Our method infers that the map in (a) is almost fully explored and stops the exploration, reducing the total exploration time.}
\label{fig:exp/progressionisbad}
\vspace{-0.5cm}
\end{figure}


In some operational contexts, such as Urban Search and Rescue \cite{liu2013robotic}, the exploration process stops after a given (usually not-so-long) amount of time because the application requires to quickly provide (possibly incomplete) information about the environment to the human rescuers. In other applications, the exploration is stopped when the process is no longer acquiring relevant information, for example, when no frontiers are left above a given length \cite{placed2022enough}. A method that decides when an exploration process should stop is called a \emph{stopping criterion}. While the need for stopping criteria is recognized, identifying robust stopping criteria is still an open challenge \cite{cadena2016}, with limited research in the field \cite{placed2022enough}.

The importance of having a good stopping criterion is paramount. Indeed, the later stages of a typical exploration process often provide little contribution to the completion of the final map, at a high cost.  As noted in \cite{ericson2021understanding}, in some settings, up to 71\% of the total exploration time is spent covering the last 10\% of the area (Fig.~\ref{fig:exp/progressionisbad}).
Moreover, late--stages of the exploration process can cause a degradation of the mapping performance, leading to map inconsistencies and unrecoverable states that invalidate the entire mapping process  \cite{placed2022enough}.

In this paper, we present a method that evaluates the partial map of an environment and estimates the advancement of the exploration process. Our method processes partial 2D grid maps obtained during exploration as images, using a Convolutional Neural Network (CNN). We exploit the fact that assessing the degree of completeness of a map is a task well suited for CNNs, as maps of indoor environments exhibit very peculiar patterns that allow humans to correctly assess -- in most cases -- whether there is still some potentially large area to explore or not. For example, looking at the partial map of Fig.~\ref{fig:map30percent}, it is intuitive to say that it represents the environment almost completely.
Our network, which is trained on an extensive dataset of maps obtained during the exploration of indoor environments, is used to: 
\begin{itemize}
    \item defining a stopping criterion that determines whether a partial map under scrutiny can be considered as enough explored or not,
    \item estimating the status of the exploration process as the percentage of the total area explored by the robot (without needing the knowledge of the whole environment), and 
    \item identifying the areas of the map that are of interest (if any) to progress the exploration.
\end{itemize}

Results show that employing our method can significantly reduce the exploration time while ensuring a good level of completeness of the map.
As our method is based on 2D grid maps, it is agnostic of the exploration strategy employed to build the map and can be used to provide additional awareness to any exploration strategy.
At the same time, we do not require any previous knowledge or priors on the environment that is explored.

\section{Related Work}\label{sec:related work}


While the generic goal of an autonomous mobile robot exploring initially unknown (static) environments is to obtain complete maps of the environments, there are often additional desiderata to address, such as maximizing the explored area within a time budget \cite{amigoni2013evaluating} or maximizing the quality of the reconstructed map \cite{placed2023survey}. In this context, the main (and most studied) issue a robot faces is to decide which is the next location to reach to progress the exploration  \cite{lluvia2021active,stachniss2005mobile}.
Over the years, different approaches to exploration have been proposed \cite{stachniss2009robotic}. A common one is to identify a few candidate locations at the boundaries between the explored and unexplored portions of the environment, and to select the most promising location to reach according to some exploration strategy. These boundaries are defined as \emph{frontiers}, and their ranking is commonly based on their expected \emph{information gain}, i.e., an estimate of the amount of knowledge to be acquired by the robot when visiting the locations, and on their \textit{distance} from the current location of the robot \cite{yamauchi1997frontier,lluvia2021active, amigoni2017multirobot}. 

Some works highlight that deciding when to stop exploration is a particularly relevant, yet under-investigated and still open, problem \cite{cadena2016,placed2022enough,placed2023survey}. 
In \cite{placed2022enough} an overview of the stopping criteria that are commonly used to end exploration is presented, as well as a discussion of their limitations. 
A straightforward criterion is to stop the exploration process when a time budget is reached \cite{leung2008active}; however, the time budget should be tuned for each environment and exploration strategy, and there is no guarantee that the map obtained by the robot when the time budget expires is complete.

Other works end the exploration when no candidate location is left \cite{yamauchi1999integrating}.
A positive aspect of this stopping criterion is that it guarantees that the map is complete; on the negative side, this completeness is obtained at the expense of a long exploration time, where most of the time is spent reaching far--away uninteresting locations that have been left behind in the initial part of exploration \cite{ericson2021understanding,RAS2020}. 
This issue is often the result of fast exploration strategies that prioritize reaching frontiers with large information gains and, consequently, encourage aggressive choices to quickly explore unobserved parts of the environments \cite{luperto2021exploration,katsumata2022map,shrestha2019learned,RAS2020}.
As pointed out in \cite{ericson2021understanding}, these strategies allow a robot to make greedy choices to map most of the environment in a short time but, from that moment on, the robot can spend up to $70\%$ of the total time in exploring marginal details of the environment that have been left behind, but that constitute about $10\%$ of the area of the environment. An example is shown in Fig.~\ref{fig:exp/progressionisbad}.


Some works assess the percentage of the total explored area to stop the exploration \cite{amigoni2013evaluating}. However, this criterion requires either the a priori knowledge of the size of the map or a way to infer it. Moreover, the fact that most of the area is already explored does not guarantee that all the areas of interest in the environment have been mapped. In \cite{bircher2018receding}, the authors apply a strategy based on rapidly-exploring random trees (RRT) to exploration, using the size of the RRT as a stopping criterion.
An alternative approach is to compare maps acquired at different times (e.g. after 10 minutes) and measure changes in the map\cite{amigoni2018improving}. When the two maps are identical, the robot did not observe any new areas during the interval, and the exploration can be stopped. However, this stopping criterion does not ensure the successful completion of the exploration task, in particular when the robot travels between two far--away frontiers.

Other works, like \cite{stachniss2003exploring}, are based on Information Theory and consider the map entropy as a stopping criterion: if the map entropy is below a given threshold, or if the map entropy increases without adding new area, the exploration can be stopped. A similar approach is applied in \cite{ghaffari2019sampling}, where entropy saturation in the current map is considered as a stopping criterion. In \cite{stachniss2005mobile}, these methods are used to stop the exploration process and mapping in dynamic environments.
However, also these techniques do not guarantee that the exploration task is really completed, and often require additional knowledge about the environment.

The work of \cite{placed2022enough} proposed to use the Theory of Optimal Experimental Design (TOED) -- which estimates the utility of a variable by a statistical analysis of its variance -- to overcome some of the above limitations, defining a stopping criterion that considers the completion of the task assigned to the robot. Authors in \cite{placed2022enough} use a D-optimality criterion, a statistical measure of the map utility, to stop the exploration when no new knowledge is added to the map. However, such a stopping criterion is intertwined with an exploration process that uses TOED, making the approach difficult to generalize to other exploration strategies, for example, those that are frontier-based. 

Very recently, the work of \cite{ericson24} used a DNN, called \emph{Floorist}, to predict unseen walls from partial maps to improve exploration performance. Differently from us, such a network is not used to infer knowledge about the status of the exploration run.
To the best of our knowledge, we are the first to leverage the visual representation of the current explored map to assess whether the remaining unexplored part of the environment is worth exploring or not. We address this decision by training a deep CNN in a supervised manner on a dataset of images generated during simulated robot explorations. The resulting solution is very practical, as it does not need any previous knowledge of the environment being explored, it is not task--dependent, and it is agnostic of the exploration strategy employed to build the maps.

\section{The Problem}
\label{sec:proposed solution}


In this paper, we propose a method that evaluates the current status of the exploration process performed by an autonomous mobile robot in an initially unknown environment. We consider a frontier-based exploration strategy \cite{yamauchi1997frontier}. However, our method relies only on an image depicting the current grid map built by the robot and could be used with any exploration strategy. Our approach consists of two aspects:
\begin{itemize} 
\item[($a$)] Classifying a partial map as sufficient (or not) to represent the environment, thus defining the stopping criterion. 
\item[($b$)] Estimating the portion of the area of an environment that is represented in the current partial map. 
\end{itemize}
Our solution can significantly improve the awareness of the robot about the map during exploration. While ($a$) can be used to stop the exploration earlier than conventional methods, thus saving time (Section \ref{sec:stopcri}), ($b$) can be used by a high-level decision-making process of single- and multi-robot systems, to plan the robot activity.

We train a deep convolutional neural network for each of the two tasks ($a$) and ($b$). The stopping criterion ($a$) is modeled as a visual recognition problem where the classification task is to distinguish between maps that are \texttt{not-explored} yet from those that are sufficiently \texttt{explored}. 
The first class contains maps that a human (who knows the full map of the environment) would consider still to be explored, typically because of missing areas that are of interest like one or more rooms that have not been mapped. The second class contains maps that a human would consider complete enough to represent the full environment. In this second case, some small parts of the environments (like corners) may not be entirely mapped yet, but such parts are not so relevant. An example of such an \texttt{explored} map can be seen in Fig.~\ref{fig:map30percent}. 
The second task ($b$), which is estimating the portion of the area already explored, is modeled as a regression problem using the CNN backbone adopted for solving the first task. 

The input for our method is a partial grid map $M_{t}$ built up to time $t$, where each cell can have three possible values: free (white), obstacle (black), or unknown (grey). In this sense, our method can be applied to any grid map $M_{t}$ regardless of how it has been obtained.

Formally, given an autonomous mobile robot that explores an initially unknown environment, we indicate with $A_{t}\in [0,1]$ the actual portion of the area that is covered in the current map $M_{t}$ at time $t$, and with $\hat{A}_{t}\in [0,1]$ the predicted portion of the explored area that is returned by our method. To formulate our problem, consider an exploration process that takes $T$ time steps to obtain a complete map $M_{T} = M$ (namely, $M_{T}$ is a map for which $A_{T} =1$). An oracle with full knowledge of the complete map $M$ could stop the exploration run at time $\bar{t} \leq T$ when all the relevant knowledge about the environment is contained in the map $M_{\bar{t}}$, namely when $M_{\bar{t}} \approx M$.
Our method, without knowing $M$, defines a stopping criterion that estimates the timestamp $\hat{t} \leq T$ when the exploration can be interrupted and for which $M_{\hat{t}} \approx M$. Note that different partial maps $M_{t}$ with the same portion of the explored area $A_{t}$ can be considered as adequately representing the whole environment or not according to how the unexplored portion of the area is distributed across the space (e.g., spread in many corners or concentrated in an unexplored room).

To solve the task ($a$) above, the stopping criterion, our method attempts to identify the first map classified as \texttt{explored} to minimize the normalized time error $err_{t} = |\bar{t} -\hat{t}|/T $.
Note that our goal is to avoid cases where the stopping criterion is triggered too soon, namely when $ \hat{t} < \bar{t}$, to avoid missing some interesting parts of the map. At the same time, triggering the stopping criterion at a time $ \hat{t} > \bar{t}$ is less critical, as it only reduces the amount of time saved without affecting relevant knowledge.

To solve the second task ($b$), explored area estimation, the goal is to reduce the area estimation error $err_{A} = |A_{t}-\hat{A}_{t} |$.




\section{The Proposed Solution}

\subsection{Input Definition and CNN Models}
\label{sec:model}

Our main intuition consists in casting the assessment of the status of the exploration in a visual recognition problem. There are two arguments supporting our choice. First of all, maps of indoor environments exhibit very peculiar patterns that allow humans to correctly assess -- in most cases -- whether there is still some potentially large area to explore or not, which is the case when the unexplored areas are little, sparse, or in regions where it is not possible to have another room (e.g., considering the adjacent explored regions). Second, tackling the problem from a visual recognition perspective allows us to leverage deep learning models, and in particular Convolutional Neural Networks (CNNs), which are very powerful also considering pre-trained architectures. 

The first step of our solution consists of creating an image $I_t$ associated with a partial map $M_t$. This is rather straightforward as the maps $M_t$ are already defined over a grid and thus can be easily centred, scaled, and resized into a $500\times500$ pixels grayscale image. Each pixel can take only 3 different values, 0 (black) for obstacles, 1 (white) for free areas, and 0.5 (grey) for unexplored areas (see Fig.~\ref{fig:Exp/NotExp}). 
  
To solve the classification ($a$) and explored area estimation ($b$) tasks previously described, we train two different CNNs, which however share the vast majority of their layers and differ only for their output layers, as they address different tasks (classification and regression). In particular, the binary classification network $\mathcal{C}$ has as output a dense layer with two neurons (corresponding to the \texttt{explored} and \texttt{not-explored} classes) with a softmax activation function. We use a threshold $\theta$ to distinguish between the two classes. This network is trained by minimizing the binary cross-entropy as a loss function. The estimation network $\mathcal{R}$  has a single output neuron with a ReLU activation function and is trained to minimize the MSE. The feature extraction component of the network is from the EfficientNet B1 \cite{tan2019efficientnet} pre-trained on ImageNet \cite{deng2009imagenet} and when training both $\mathcal{C}$ and $\mathcal{R}$ we perform transfer learning by freezing all the parameters of the first 200 layers. We selected the EfficientNet B1 model as a good tradeoff between accuracy and computational complexity (enabling timely inference even on a CPU), even though it is possible to consider other backbone networks.


\subsection{Datasets}\label{sec:datasets}
To train and test our models, we use a set of partial maps $M_{t}$ collected at different exploration stages, each one paired with the corresponding label (\texttt{explored} or \texttt{not-explored}) and the percentage of explored area $A_{t}$. 

The partial maps are acquired incrementally during the robot explorations performed for a previous work~\cite{amigoni2018improving,luperto2021predicting}, where we collected over $20,000$ partial maps at different stages of exploration from 100 different large--scale indoor environments. All the maps in our dataset are free from clutter (e.g., objects, furniture, \ldots). Every environment is autonomously explored multiple times by a simulated robot using the method of \cite{yamauchi1997frontier}; in each run, due to simulated noise in perception, the robot makes different choices during exploration, follows different trajectories, and ultimately builds different maps. During exploration, partial maps of the environments are saved every 10 minutes. The number of runs for each environment depends on the complexity of the environment; on average, we collected $36$ runs for each environment. See \cite{amigoni2018improving,luperto2021predicting} for further details.

We automatically label each partial map $M_{t}$ as \texttt{explored} and \texttt{not-explored} using a rule--based approach, which employs the full map $M$ of the environment as a reference. 
%
%
\begin{figure}[t]
\centering
\subfloat[\texttt{not-explored} \\ $A_{t}=0.77$.\label{fig:not_exp_map}]{\includegraphics[trim={4.5cm 0cm 4.5cm 0cm},angle=90,clip,width=0.7\linewidth]{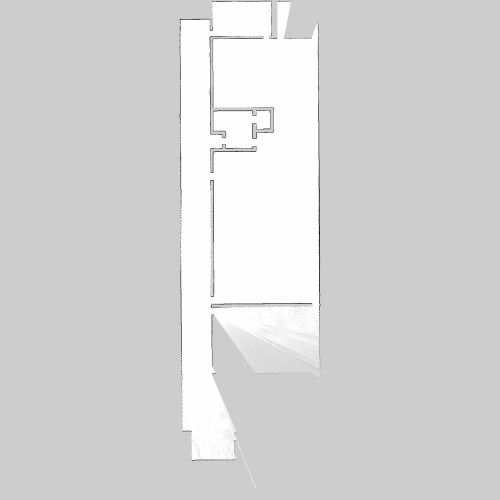}}\\
\subfloat[\texttt{explored} \\ $A_{t}=0.99$.\label{fig:exp_map}]{\includegraphics[trim={4.5cm 0cm 4.5cm 0cm},angle=90,clip,width=0.7\linewidth]{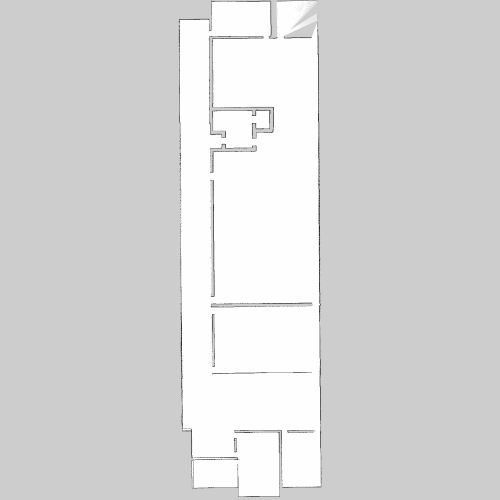}}\\
\subfloat[Complete map.\label{fig:gt_map}]{\includegraphics[trim={4.5cm 0cm 4.5cm 0cm},angle=90,clip,width=0.7\linewidth]{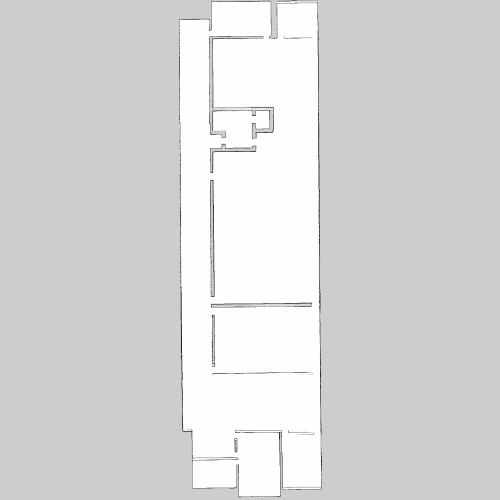}}
\caption{Examples of partial maps and labels from our dataset.}
\label{fig:Exp/NotExp}
\vspace{-0.4cm}
\end{figure}
%
 %
First, for each partial map $M_{t}$, we compute the portion $A_{t}$ of the explored area by comparing $M_{t}$ with the complete map $M$ of the corresponding environment. When $A_{t}$ is smaller than a threshold $\alpha = 0.7$, the map is safely labeled as \texttt{not-explored}. The remaining partial maps (having less than $30\%$ of the area unexplored) can be either labeled as \texttt{explored} or not depending on the spatial distribution of the unexplored pixels. More precisely, if the unexplored area is scattered in multiple small locations across the environment, as in the example of Fig.~\ref{fig:map30percent}, the map is considered \texttt{explored}. Instead, if there is a large unexplored area, like a full unexplored room, the map is labeled as \texttt{not-explored}. To distinguish between the two cases automatically during the training set preparation procedure, we cluster the unexplored cells in the partial map $M_{t}$ using DBSCAN \cite{ester1996density}. If the unexplored area corresponding to the largest cluster is greater than a threshold $\beta = \SI{1}{\square \meter}$, the map is labeled as \texttt{not-explored}; otherwise, we consider it \texttt{explored}. 
We have validated the correctness of the automated labeling process by visually inspecting its results.
%
%
Examples of maps belonging to the two classes are shown in Fig.~\ref{fig:Exp/NotExp}.
\begin{figure*}[ht]
\small
\centering
\includegraphics[width=\linewidth]{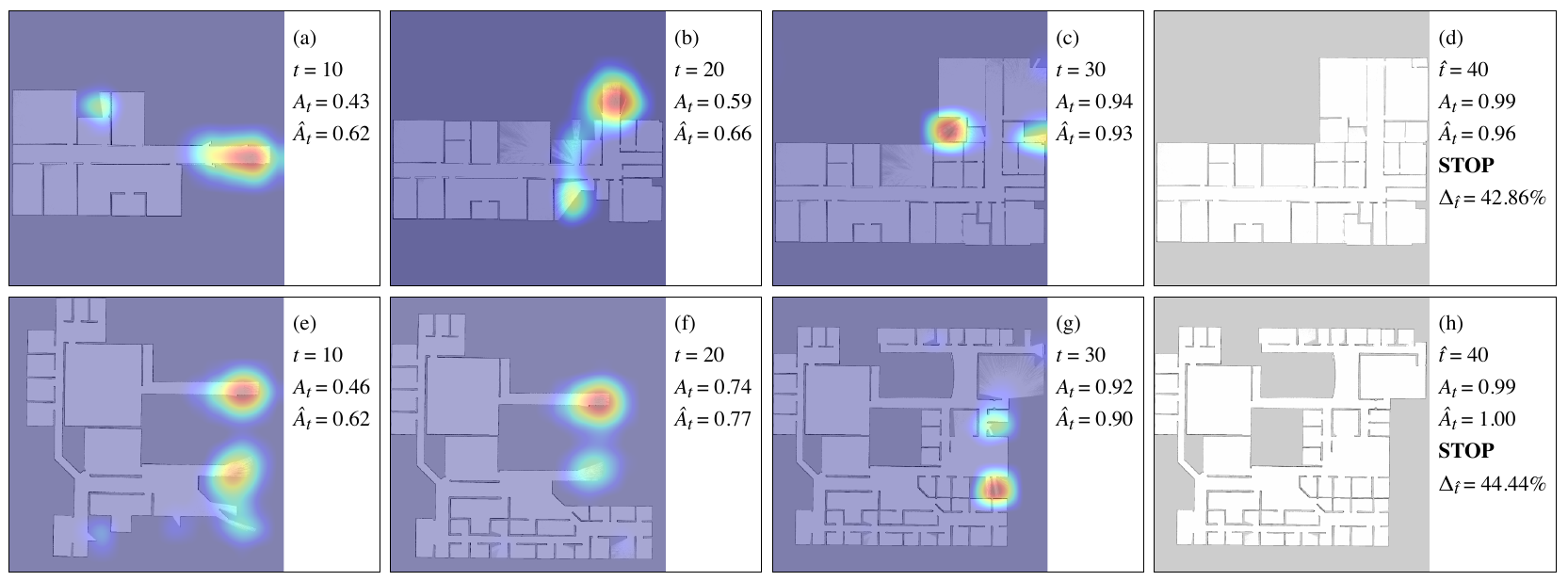}
\caption[Grad-CAM visualization examples.]{
Incremental examples of our method applied to partial maps  obtained after 10, 20, 30, and 40 minutes of exploration of two environments, from batch-wise assessment. In the environment in the first row (a)-(d), the exploration with the baseline stopping criterion ends after $T=\SI{70}{\minute}$; with our method, we can reduce such time by 30 minutes (d). In the environment in the second row (e)-(h), the exploration with the baseline stopping criterion is concluded after $T=\SI{90}{\minute}$;  we reduce such time by 50 minutes (h).
The areas highlighted in the partial maps are the locations that the neural network of our method identifies as relevant to make its prediction (red and blue denotes the most and the least relevant regions, respectively).}
\label{fig:grads_1}
\end{figure*}

\subsection{Model Training}
The environments in the dataset are split into the training, validation, and test set in a $70:15:15$ ratio. To artificially increase the size of our training and validation sets and lower the risk of overfitting, we use image augmentation by changing the orientation and zoom of the original images. 
We use a batch size of $8$ and we train the model for 24 epochs, using a learning rate of $1 \cdot e^{-4}$ and the Adam optimizer \cite{kingma2014adam}. 


\subsection{Stopping Criterion}\label{sec:stopcri}

Our stopping criterion is triggered when a partial map obtained during the exploration process is labeled as \texttt{explored} by our model. A more conservative approach can be developed when different factors are considered, for example when multiple consecutive partial maps are labeled as \texttt{explored} or $\hat{A}_{t}$ is above a threshold. 
\red{In this sense, the estimated percentage of the explored area can be exploited to inform the exploration strategy as discussed in Section~\ref{sec:intro}.}

\red{An additional benefit of our method is that we can leverage saliency maps on the trained CNN classifier to design new exploration strategies. Saliency maps, like Grad-CAM \cite{gradcam}, are explainability techniques that provide, for each input image, a heatmap that highlights which regions have primarily influenced the network to decide a specific class. Therefore, saliency maps associated with the class  \texttt{not-explored} can suggest which region to prioritize in the exploration, as shown in Fig.~\ref{fig:grads_1}.
However, the investigation of novel exploration strategies based on the returned exploration percentage and the saliency maps is left for future work.}

\section{Experimental Evaluation}
\label{.:experimental evaluation}

In this section, we first assess the performance of using our CNN $\mathcal{C}$ for classifying maps in \texttt{explored} / \texttt{not-explored} (Section \ref{sec:accuracy}), as well as the corresponding exploration time saved by adopting our stopping criterion (Section \ref{sec:sc}). We then evaluate the performance of our CNN $\mathcal{R}$ in estimating the portion of area explored (Section \ref{sec:area}).
Moreover, we show some qualitative results that give insights into how our method works (Section~\ref{sec:qualitative}), and
we discuss how our method can be applied to data obtained from real-world environments (Section~\ref{sec:real}). 

Our evaluation is performed by using a set of maps obtained by the incremental exploration of indoor environments as explained in Section \ref{sec:datasets}. The environments considered for evaluation have not been used for training and are of different sizes and shapes to represent all possible types of environments. They vary from small-scale environments with less than 10 rooms and smaller than \SI{100}{\square \meter} to large-scale environments with more than 100 rooms covering more than \SI{10000}{\square \meter}.
To assess the benefits of our approach, we run both a batch-wise (\texttt{B}) and an online (\texttt{O}) assessment. 

The \emph{batch-wise} evaluation allows us to understand the performance of the proposed solution when processing maps that are visually different from each other, thus ensuring the generalization capabilities of our solution. Batch-wise evaluation is performed over a test set containing 3009 maps obtained from 15 environments. Overall, we performed 613 robot exploration runs; during each run, every 10 minutes, a snapshot of the current map of the environment was saved. The time interval between two snapshots ensures that two consecutive maps obtained by the robot are significantly different, as can be seen in the example of Fig. \ref{fig:grads_1}, where maps obtained in two different robot runs are shown.

The \emph{online} evaluation aims at assessing the benefits of using our method during the exploration when the robot frequently decides whether to continue exploring, based on its current map. To replicate this setting, we consider maps that are collected during an exploration run at a higher frequency, namely every \SI{30}{s}. We consider 10 environments and we perform 5 exploration runs in each environment, for a total of 50, obtaining 2255 maps. Note that online assessment cannot be used for a fair evaluation of the map classification performance, as consecutive maps are very similar, due to the higher sampling frequency. Our classifier can easily distinguish that all maps obtained during the early (late) stages of explorations are \texttt{unexplored} (\texttt{explored}), resulting in an optimistic performance assessment. At the same time, batch evaluation cannot provide a precise estimate of the stopping criterion, due to the interval of time spent between the acquisition of two maps.

\red{All exploration runs of our experimental evaluation continue until a \emph{baseline} stopping criterion is met, which consists of ending the exploration at time $T$ when two partial maps, obtained at a time difference of \SI{10}{\minute}, are similar enough. We measure map similarity by the vector norm of the difference between the two images, as in \cite{amigoni2018improving}. We manually inspected all exploration runs to ensure that the baseline stopping criterion is triggered only when the environment is fully explored; if the baseline stopping criterion is triggered too early, the run is removed and another run is performed. Note that this baseline stopping criterion is more efficient than others considered in the literature. As an example, we do not require the robot to visit all of the remaining frontiers, as in \cite{yamauchi1999integrating}: when two partial maps are similar enough and the environment is considered as fully explored according to manual inspection, the baseline exploration criterion is triggered even when uninteresting frontiers are still present in the environment. We do not consider other baseline stopping criteria as they either achieve worse performance (e.g., they require a large time budget to guarantee that the full environment is mapped, as in \cite{leung2008active}) or require the robot to follow a specific exploration strategy (e.g., \cite{placed2022enough}).}

We analyze the impact of varying the threshold $\theta$ for classifying a map as explored or not (Section~\ref{sec:model}). We use a value of $\theta = 0.5$ as a reference and we also present results with a more conservative parameter value $\theta = 0.8$, which reduces the number of partial maps that are wrongly considered as \texttt{explored}.

The training and testing of our CNNs are performed on a computer with an i7 3770k CPU, 16 GB RAM, and RTX 2070 GPU. The task of evaluating the status of the exploration process from a partial map requires less than \SI{0.1}{\second} and can be performed online. Further experimental results and details are available in our online repository\footnote{\url{https://aislabunimi.github.io/explore-stop/}}.
Since partial maps $M_t$ need to be processed at a relatively low frequency, and since the EfficientNetB1 is a lightweight network, our method can be directly executed on a CPU with an inference time of less than \SI{4}{\second} \cite{tan2019efficientnet}, thus not requiring robots to be equipped with a GPU.

\begin{figure}[t]
\centering
\includegraphics[width=0.95\columnwidth]{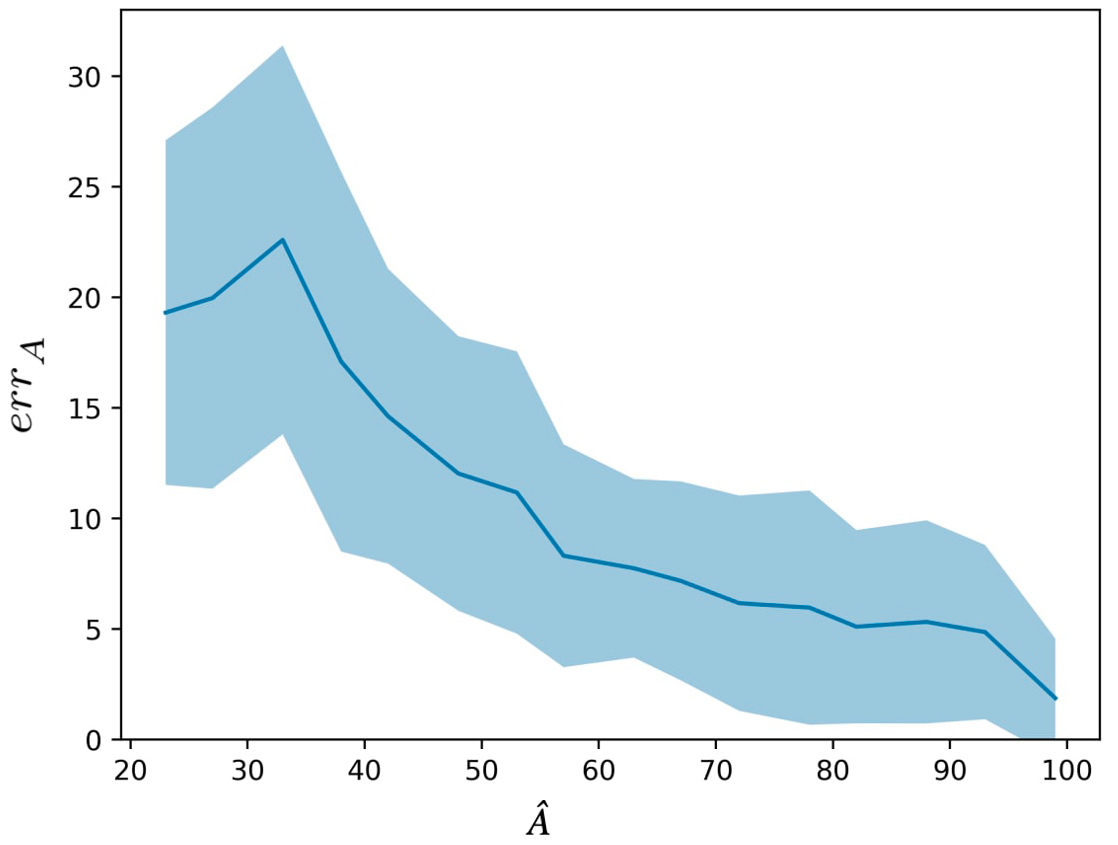}
\captionsetup{width=.95\columnwidth}
\caption{The error in predicting the percentage of explored area while the percentage of the area increases. The standard deviation is in light blue. }
\label{fig:predicted}
\end{figure}

\begin{figure*}[t]
\centering
\includegraphics[width=\linewidth]{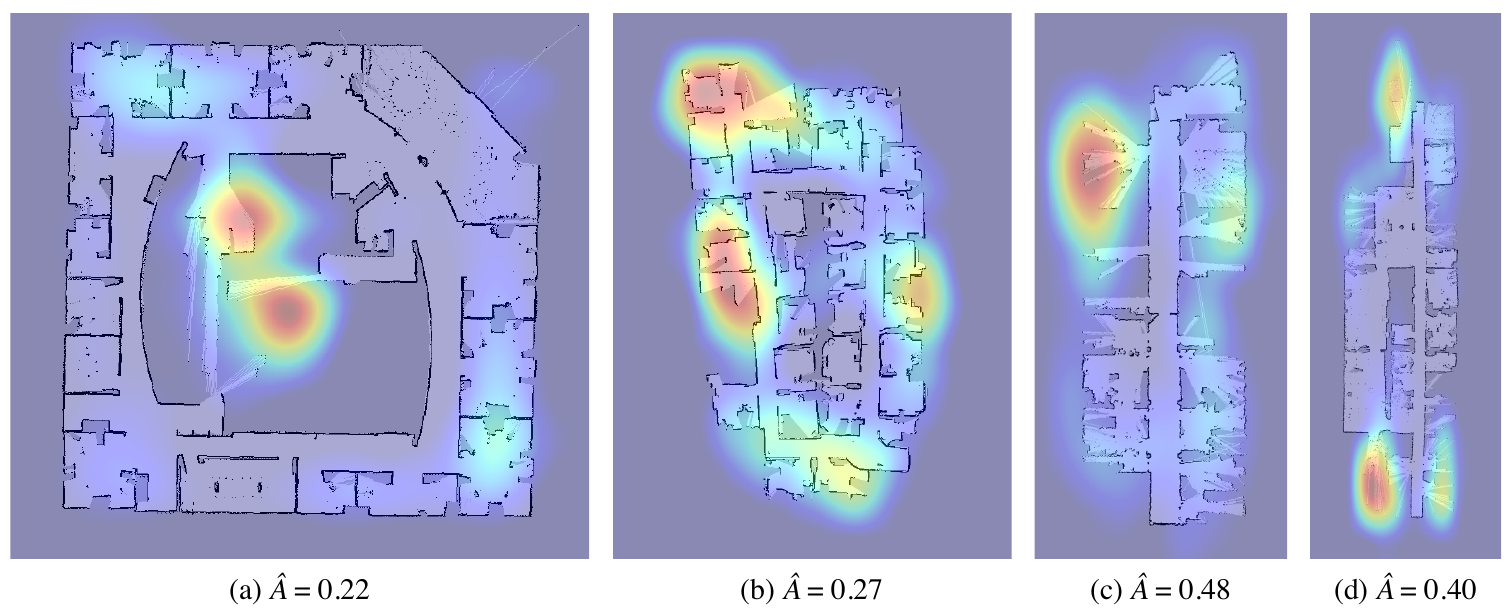}
\caption{Examples of real-world maps, all labeled as \texttt{not-explored}. 
Our method can highlight the parts relevant to explore as well as predict the explored area $\hat{A}$.}
\label{fig:Exp/real}
\end{figure*}

\subsection{Map Classification Performance}\label{sec:accuracy}

We indicate as a True Explored (True Not-Explored) or TE (TN) the percentage of test maps correctly classified as \texttt{explored} (\texttt{not-explored}), and as False Explored (False Not-Explored), or FE (FN), the classification errors. We report accuracy (A), precision (P), and recall (R) of the classification.

Table \ref{table:Conf} shows the results for both the batch and online evaluation. For the batch evaluation (\texttt{B}) with $\theta=0.5$, the accuracy A is 92.9\%, and only 1.76\% of the examples are false positives FE. Minimizing FE is particularly important as a map that is wrongly considered as \texttt{explored} (false positives, FE) can result in loss of relevant information in the map. In contrast, FNs are less harmful, as it is acceptable to consider as not finished the exploration of an actually explored map. However, in our experiments, false positives FE have a minimal impact on the performance metric; a map wrongly labeled as \texttt{explored} covers, on average, 2.51\% less area than the first truly \texttt{explored} map obtained in the same run. 
Increasing $\theta$ to $0.8$ further reduces FE errors, at the cost of slightly reducing the value of TE. To evaluate the effects, we manually inspect all 21 false positives FE with $\theta=0.8$. \red{They come from 15 of the 613 runs and refer to 7 of the 15 environments, indicating that some features of the environment can lead to an FE. In fact, in all cases, the FE is caused by an entire room that is missing from the map, without any single beam of the robot's lidar entering its door. Similar remarks can be made by looking at the metrics computed on the maps used for the online evaluation \texttt{O}.}

\begin{table}[]
\resizebox{\linewidth}{!}{
\begin{tabular}{|l|l|l|l|l|l|l|l|l|}
\hline
                          & $\theta$ & TE (\%)   & FE (\%) & TN (\%)   & FN (\%)  & A    & P    & R    \\ \hline\hline
\multirow{2}{*}{B} & 0.5     & 1472 (48.91) & 53 (1.76) & 1324 (44.00) & 160 (5.31) & 0.92 & 0.97 & 0.91 \\ \cline{2-9} 
                         & 0.8     & 1376 (45.73) & 21 (0.70) & 1352 (44.93) & 260 (8.60) & 0.91 & 0.99 & 0.84 \\ \hline \hline
\multirow{2}{*}{O}  & 0.5     & 970 (43.00)  & 48 (2.12) & 1116 (49.49) & 121 (5.37) & 0.93 & 0.95 & 0.89 \\ \cline{2-9} 
                         & 0.8     & 926 (41.06) & 28 (1.24) & 1136 (50.38) & 165 (7.32) & 0.91 & 0.97 & 0.85 \\ \hline
\end{tabular}}
\caption{Map classification as \texttt{(not)-explored} during batch (\texttt{B}) or online (\texttt{O}) evaluation. }
\label{table:Conf}
\end{table}

\subsection{Stopping Criterion Evaluation}\label{sec:sc}
To assess the effectiveness of our stopping criterion (with $\theta=0.5$), which stops the exploration when a partial map $M_{t}$ is labeled as \texttt{explored}, we compute the percentage of the time saved compared to an  exploration ending with the baseline stopping criterion $\Delta \hat{t} =| T - \hat{t}| / T  $, as well as the average of $\Delta\hat{t}$ across all runs in all environments.
Table \ref{table:all} (left) shows the results of the batch evaluation, averaged across all runs for each test environment. Our stopping criterion can reduce the total exploration time (against the baseline) by 31.62\%, on average. Note how our performance is close to those of an ideal stopping criterion (as defined in Section~\ref{sec:proposed solution}), which for our test environments of the batch evaluation has an average time saved of $\Delta \bar{t} = | T - \bar{t}| / T  = 33.46\%$. As a consequence, $err_{t}$ of Section~\ref{sec:proposed solution} is small, under $2\%$.

While already promising, we emphasize that the batch evaluation reported in Table \ref{table:all} (left) provides a rough estimation of the actual performance of our stopping criterion, as the test dataset for \texttt{B} contains maps that are saved every 10 minutes during a run. \red{Indeed, during an exploration run, it is likely that the stopping criterion should be triggered at a time that lies within the 10-minute interval.}


Results on the online settings, where maps are saved every 30 seconds during each run, provide a more realistic evaluation of the speed up enabled by our stopping criterion, which can save about $40\% $ of exploration time, and are reported in Table \ref{table:all} (right).  This significant decrease in the exploration time comes at the small cost of stopping the exploration when less than 1\% of the total area of the environment is still to be mapped.
\red{Remarkably, our stopping criterion is triggered in all runs at the same time as the ideal stopping criterion $\bar{t}$, thus having a $err_{t} = |\bar{t} -\hat{t}|/T = 0$.
A more conservative $\theta=0.8$ result in $\Delta \bar{t} = 35\%$ and $err_{t}=1.92\%$. }



\begin{table}[t]
\small
\centering
\begin{adjustbox}{max width=\columnwidth}
\centering
\begin{tabular}{|c|c|c|c|}
\hline
\textbf{envs} & \textbf{\# runs} & \textbf{$\Delta \hat{t}$} &  \textbf{$err_{A}$} \\
\hline\hline
b$_1$ & 41 & 43.54\% & 4.57\% \\
\hline
b$_2$ & 36 & 19.57\% & 4.40\% \\
\hline
b$_3$ & 43 & 41.23\% & 5.04\%\\
\hline
b$_4$ & 44 & 29.09\% & 3.70\% \\
\hline
b$_5$ & 39 & 48.71\% & 3.60\% \\
\hline
b$_6$ & 30 & 24.16\% & 5.90\%\\
\hline
b$_7$ & 40 & 24.40\% & 4.40\%\\
\hline
b$_8$ & 47 & 31.48\% & 4.50\%\\
\hline
b$_9$ & 39 & 50.00\%& 5.30\% \\
\hline
b$_{10}$ & 46 & 14.40\% & 7.03\%\\
\hline
b$_{11}$ & 52 & 21.20\% & 7.80\%\\
\hline
b$_{12}$ & 36 & 27.00\% & 2.20\%\\
\hline
b$_{13}$ & 34 & 23.77\% & 5.40\%\\
\hline
b$_{14}$ & 56 & 48.21\% & 2.20\% \\
\hline
b$_{15}$ & 30 & 27.63\% & 4.10\%\\
\hline\hline
all & 613 & 31.62\% & 4.68\%\\
\hline
\end{tabular}
\quad

\begin{tabular}{|c|c|c|}
\hline
\textbf{envs} & \textbf{\# runs} & \textbf{$\Delta \hat{t}$}\\
\hline\hline
o$_{1}$ & 5 & 29.8\%\\
\hline
o$_{2}$ & 5 & 59\%\\
\hline
o$_{3}$ & 5 & 40.69\%\\
\hline
o$_{4}$ & 5 & 31.3\%\\
\hline
o$_{5}$ & 5 & 37\%\\
\hline
o$_{6}$ & 5 & 34\%\\
\hline
o$_{7}$ & 5 & 36\%\\
\hline
o$_{8}$  & 5 & 26.88\%\\
\hline
o$_{9}$ & 5 & 36.83\%\\
\hline
o$_{10}$ & 5 & 41.17\%\\
\hline\hline
all & 50 & 37.37\%\\
\hline
\end{tabular}
\end{adjustbox}

\caption{Saved time with our stopping criterion and error in predicting the explored area for each environment of the batch (\texttt{B}, left) and online (\texttt{O}, right) evaluation.}
\label{table:all}
\end{table}

\subsection{Map Completion Percentage Prediction}\label{sec:area}

The fourth column of Table \ref{table:all} (left) reports the average error $err_{A}$ in estimating the area covered by the map $M_{t}$ (Section~\ref{sec:proposed solution}). Overall, we observe an average value of $err_{A} = 4.68\%$, which is low and stable across all 15 test environments. To enable a fair comparison,  we report only the data collected with the batch evaluation as, during the late stages of the online evaluation, we typically have multiple almost-complete maps which are correctly identified by our method, thus overestimating its performance. 

Fig.~\ref{fig:predicted} shows how the average  $err_{A}$ progresses during an exploration run. While at the beginning of the exploration, the error is higher (as the portion of the map known to the robot is too little to make an accurate estimation), $err_{A}$ becomes stable and under $10\%$ when half or more of the environment is explored.
Overall, our method can provide a reliable estimate of the amount of the explored area represented in a map. 

\subsection{Qualitative Evaluation}\label{sec:qualitative}

\red{Fig.~\ref{fig:grads_1} shows two examples of maps that are incrementally obtained during exploration runs from the batch evaluation \texttt{B}, as processed by our framework. Figs. \ref{fig:grads_1}(a-c) and Figs. \ref{fig:grads_1}(e-g) show how saliency maps from Grad-CAM can identify relevant areas to be explored.} Figs. \ref{fig:grads_1}d and \ref{fig:grads_1}h show the maps when the stopping criterion is triggered. The maps are correctly considered as complete, 30 and 50 minutes before the actual end of the exploration runs when \red{using the baseline stopping criterion, thus saving $\approx 40\%$ of the time.}
Also, our method can robustly predict the percentage of the explored area, increasing the accuracy of the estimate as the exploration progresses.

\subsection{Real-World Maps}\label{sec:real}

As our method uses simulated and empty maps for training, we run the following test to assess whether some domain adaptation technique is needed when adopting it in real-world maps acquired in cluttered environments. More precisely, we test our solution on 64 maps of real-world environments. 54 maps are obtained during the incremental mapping process of 3 real-world environments, as recorded in the publicly-available dataset of \cite{radish}. The remaining 10 maps are obtained during the frontier-based exploration process of a large-scale real-world environment from Matterport \cite{Matterport}. We label all the 64 maps as \texttt{not-explored} by a visual inspection, as in all these maps some relevant portions of the environments are not yet mapped by the robot.

We remark that these maps are significantly different from those used to train our model, which are obtained in empty environments without clutter. Despite this domain shift, our model shows robust performance by correctly classifying all the maps as \texttt{not-explored}. Fig.~\ref{fig:Exp/real} shows Grad-CAMs for four examples, highlighting how in all cases our model can accurately identify what is still left to explore, while also estimating the explored area $\hat{A}$.
As we do not know the actual area $A$ of these environments, we cannot compute $err_{A}$.  
Still, we can observe how in all 64 maps our method correctly highlights the parts of the environments that are relevant to explore and provides a plausible monotonically increasing estimate of $\hat{A}$.

\section{Conclusion}
\label{sec:conclusions}

In this work, we have presented a method to increase the awareness of an autonomous mobile robot during the process of exploration for map building of unknown environments. 
Our method, given the partial grid map of an environment, provides a reliable estimate of the explored area and predicts if the exploration run can be stopped (or not) because the map contains all the relevant knowledge about the environment (or not).
Results show that we reduce the total time required for exploration up to $40\%$.

Future work involves using the output of our method to inform exploration strategies, improving the quality of the resulting map in tasks like Active SLAM, and using the estimate $\hat{A}$ to support the exploration process with a time budget, e.g., by switching between different strategies \cite{amigoni2017online}. 
In general, our method can be used to define different stopping criteria that can be used in conjunction with different 
desiderata of the exploration process. 
We will also investigate how a similar pipeline can be applied to other spatial representation methods, such as 3D maps and Truncated Signed Distance Fields maps.
\bibliographystyle{IEEEtran}
\bibliography{citations}

\end{document}